\pdfoutput=1

\documentclass[11pt]{article}

\usepackage[preprint]{acl}
\usepackage{times}
\usepackage{latexsym}
\usepackage{amsmath}
\usepackage{amsfonts}
\usepackage{enumitem}
\usepackage{float}
\usepackage{tcolorbox}
\usepackage{booktabs}
\usepackage{makecell}
\usepackage[table]{xcolor}
\usepackage{colortbl}
\usepackage{amssymb}
\usepackage{multirow}
\usepackage{tabularx}
\usepackage[normalem]{ulem}

\usepackage{arydshln}

\usepackage{mathtools}
\definecolor{green1}{HTML}{198038}
\definecolor{red1}{HTML}{DA1E28}
\definecolor{softblue}{RGB}{236,243,252}

\usepackage[T1]{fontenc}

\usepackage[utf8]{inputenc}

\usepackage{microtype}

\usepackage{inconsolata}

\usepackage{graphicx}
\usepackage{xspace}

\title{Valid Survey Simulations with Limited Human Data:\\
The Roles of Prompting, Fine-Tuning, and Rectification}

\author{
  \textbf{Stefan Krsteski\textsuperscript{1}\thanks{Equal contribution.}}, 
  \textbf{Giuseppe Russo\textsuperscript{1,2}\footnotemark[1]}, 
  \textbf{Serina Chang\textsuperscript{3}}, 
  \textbf{Robert West\textsuperscript{1}}, 
  \textbf{Kristina Gligorić\textsuperscript{4}} \\
  \textsuperscript{1}EPFL \\
  \textsuperscript{2}Stanford University \\
  \textsuperscript{3}University of California, Berkeley \\
  \textsuperscript{4}Johns Hopkins University
}

\definecolor{green1}{HTML}{2EBC99}
\definecolor{red1}{HTML}{D04F4F}
\definecolor{PurpleHighlight}{HTML}{9B59B6}
\definecolor{BlueHighlight}{HTML}{0082FB}
\definecolor{OrangeHighlight}{HTML}{F18F01}
\definecolor{GreenHighlight}{HTML}{74AA9C}

\begin{document}
\maketitle
\begin{abstract}
Surveys provide valuable insights into public opinion and behavior, but their execution is costly and slow. Large language models (LLMs) have been proposed as a scalable, low-cost substitute for human respondents, but their outputs are often biased and yield invalid estimates. We study the interplay between synthesis methods that use LLMs to generate survey responses and rectification methods that debias population estimates, and explore how human responses are best allocated between them. Using two panel surveys with questions on nutrition, politics, and economics, we find that synthesis alone introduces substantial bias (24–86\%), whereas combining it with rectification reduces bias below 5\% and increases effective sample size by up to 14\%. Overall, we challenge the common practice of using all human responses for fine-tuning, showing that under a fixed budget, allocating most to rectification results in far more effective estimation.

\end{abstract}



\section{Introduction}
Self-reported surveys are the gold standard for capturing how people think, feel, and behave across domains such as public policy, economics, and health. However, they are costly, time-consuming, and logistically complex \cite{groves2011survey}. Recent works at the intersection of survey research and natural language processing have explored using LLMs as proxies for human respondents \cite{gao2024large,lira2022large, argyle2023out, anthis2025llm, bail2024can}.

Despite their potential~\cite{manning2024automated,shah2025learning}, LLMs are not reliable out-of-the-box as survey respondents~\cite{gao2024large}. Empirical studies document demographic and positional biases~\cite{cheng2023marked,wang2025large}, sensitivity to prompt wording, lexical features, and option order~\cite{atreja2025s,gligoric-etal-2024-nlp}, desirability bias~\cite{sharma2023towards,cheng2025social}, and hallucinations or self-contradictions \cite{tjuatja2024llms,dominguez2024questioning,pezeshkpour2023large,huang2025survey}. Naïve use can therefore distort population estimates. Training-time adaptations such as fine-tuning on survey responses demand extensive human annotation and remain vulnerable to domain shift, whereas inference-time techniques like demographic prompting or persona-based generation are highly prompt-sensitive \cite{sun2025sociodemographic}. 

Methods such as \textit{Prediction-Powered-Inference} (PPI) and \textit{Design-based Supervised Learning} (DSL)~\cite{angelopoulos2023prediction,egami2023using} have been proposed as a post-hoc correction approach, but they have not been rigorously evaluated for large-scale survey simulation or in combination with training- and inference-time adaptations often used in practice. Moreover, these approaches guarantee validity only for corrected estimates, not for the generated responses themselves. Ensuring that generations are unbiased remains important when follow-up questions are posed \cite{wang2024llmcheckup,shaikh-etal-2024-grounding} or when other quantities beyond the corrected estimate need to be inferred. 

Consequently, the effectiveness of training-time, inference-time, and post-hoc methods for valid LLM-based survey simulation is still unclear, as are their potential interactions. A priori, the optimal allocation of limited human data across these methods is not evident. For example, dedicating all data to fine-tuning precludes effective post-hoc correction, while allocating none may compromise the quality of generated responses. Nonetheless, these interactions remain uncharacterized and no guidelines exist to date.

\begin{figure*}[ht!]
    \centering
    \includegraphics[width=\linewidth]{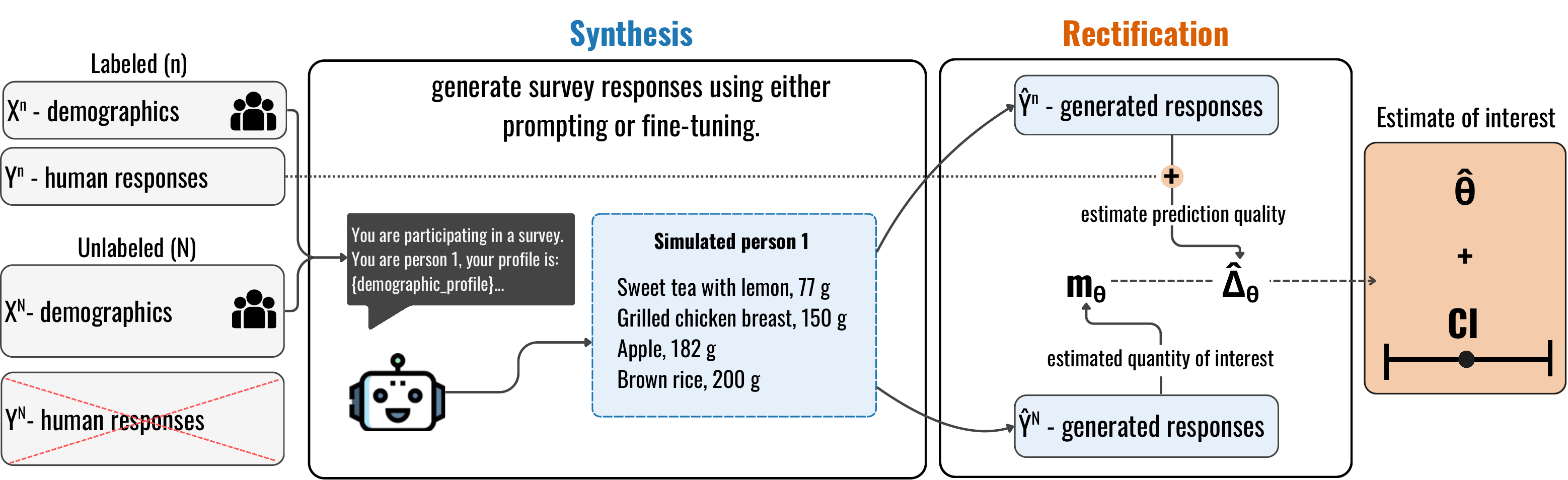}
    \caption{\textbf{Evaluation setup: Overview of synthesis and rectification.} 
    Given a small human dataset $(X^n, Y^n)$ and a disjoint, large demographic only dataset $X^N$, 
    \texttt{Synthesis} produces responses $\hat{Y}^n, \hat{Y}^N$ using either prompting or fine-tuning. 
    \texttt{Rectification} then combines model predictions $\hat{Y}^n$ with human responses $Y^n$ to compute a correction term. Then, this term is combined with $\hat{Y}^N$ to produce a final estimate $\hat{\theta}$ of the target $\theta^*$, with corresponding confidence intervals.}
    \label{fig:flowchart}
\end{figure*}

To address these gaps, we conduct an evaluation examining supervised fine-tuning, persona-guided prompting, and rectification methods, and how to allocate gold-standard human responses for maximizing performance. Our evaluation is carried out on two large-scale surveys: the NHANES dietary recall survey \cite{national2017national} and the American Trends Panel (ATP) \cite{pew_atp}. Studying longitudinal data allows us to address the most general setting, from which simpler survey designs can be derived. Through this setup, we contribute a head-to-head, format-agnostic, budget-aware study of training-time (fine-tuning), inference-time (prompting), and post-hoc rectification methods for survey simulation. Our contributions are the following:

\begin{enumerate}
\item \textbf{A rigorous evaluation} of four common synthesis strategies combined with two rectification methods, grounded in two large-scale surveys with focal tasks in food, politics, and economics (\textsection\ref{sec:experiments}, \textsection\ref{sec:results1}).
\item \textbf{Quantitative insights} showing that population and subpopulation estimates can be corrected to within 5\% bias using as few as 100 responses ($\approx$1\% of a full survey) (\textsection\ref{sec:results2}).
\item \textbf{Evidence-based guidelines} for allocating responses across synthesis and rectification (Table~\ref{tab:guidelines}). For example, while dedicating all data to fine-tuning may seem natural, we show that under a fixed budget, allocating the majority of responses to rectification yields the best bias–variance trade-off.

\end{enumerate}
All code and data are released to support development of new methods and benchmarks\footnote{Code is available at: \href{https://anonymous.4open.science/r/synrec-618E}{anonymous.4open.science/synrec}}.



\section{Related work}

\paragraph{Training-time adaptation.}
LLMs have been proposed as proxies for human respondents, enabling low-cost, large-scale survey simulation in domains such as public opinion, voting\cite{argyle2023out,santurkar2023whose}. A common way to improve domain performance is fine-tuning. Recent work aligns response distributions by minimizing a forward KL divergence between model probabilities and human distributions for a given (question, subgroup) pair \cite{suh2025language, cao2025specializing, davidson2024self, russo2020control}. First-token probability alignment is typical for multiple-choice questions (MCQs). Although LoRA and other PEFT techniques reduce computation, fine-tuning still requires unpredictable amounts of gold-standard data and meticulous preprocessing. \cite{hu2022lora,suh2025language,wang2025parameter}. Moreover, survey-specific artifacts require careful format alignment \cite{dominguez2024questioning, wang2024my}. This raises the question of where scarce human responses are most valuable, but existing studies provide no principled answer. Our evaluation 
aims to address this gap.

\paragraph{Inference-time adaptation.}
Prompting and in-context learning adapt the model at inference without parameter updates, making them a popular choice for survey simulation \cite{brown2020language, latona2024ai, russo2025does}. Early work such as ``silicon sampling'' shows that sociodemographic conditioning can elicit responses with high population-level fidelity to humans \cite{argyle2023out,santurkar2023whose, sun2024random}. Later methods introduce synthetic personas and steering~\cite{hu2024quantifying}, including steering vectors \cite{kim2025linear, russo2025pluralistic}, soft-prompt control \cite{li2023steerability,li2025llm}, and probabilistic mixtures of persona-conditioned agents \cite{bui2025mixture}.
Despite this progress, evaluations report prompt sensitivity, question- and option-order artifacts, and demographic skew, casting doubt on reliability for population-level inference \cite{dominguez2024questioning,tjuatja2024llms,geng2024large}. A related concern is format-induced bias: many studies focus on MCQ/Likert formats, whereas prior work shows that responses under constrained generation can diverge from open-ended ones \cite{rottger2024political,zhang2025simulation}. If minor setup changes (e.g., prompt or model) lead to substantially different responses, the validity of conclusions from such simulations is questionable. In this work we show that incorporating even a small amount of human responses via rectification can mitigate such biases.

\paragraph{Post-hoc adaptation.}
A complementary strategy bypasses model editing, treating the LLM as a powerful but biased black-box predictor whose outputs are statistically corrected. Frameworks such as PPI \cite{angelopoulos2023prediction} and confidence-driven inference (CDI) \cite{gligoric2024can} provide finite-sample–valid estimators and confidence intervals by combining abundant model predictions with a small human subset. Although previously applied to annotation tasks~\cite{fannarratives,calderon2025alternative,rister2025correcting}, this paradigm fits survey workflows where instrument design is costly, but collecting a small subsample of human answers is easy. In social-science methodology, design-based semi-supervised learning (DSL) combines model predictions with a small human sample via a doubly robust estimator \cite{egami2023using}. Similarly, the mixed-subjects design incorporates LLM predictions as additional observations alongside human responses \cite{broska2025mixed}. However, existing validations focus on controlled behavioral tasks rather than large-scale survey simulation, and they do not examine interactions with different synthesis strategies or data-allocation trade-offs. Here, we provide practical guidance on how to get the best of both worlds, simultaneously debiasing synthesis models and population-level estimates. We fill this gap with a multi-dataset evaluation that pairs several synthesis choices (prompting/personas and fine-tuning) with post-hoc correction across both MCQ and open-ended questions. 

\section{Methodology}\label{sec:framework}

The task of interest is producing accurate population-level estimates that reflect human survey responses. To this end, we evaluate strategies that synthesize responses conditioned on demographics, correct estimates using a small subset of human responses, and combine both to balance their strengths. 

\paragraph{Problem formalization.}
We frame the problem in the general setting of panel surveys, from which cross-sectional designs arise as a special case. Thus, the formulation leverages histories when available and reduces to the cross-sectional case when not. A panel survey is a sequence of responses collected from the same \(N\) individuals over \(T\) discrete time points. A single time point \(t\) (a ``wave'') may represent, for example, a monthly opinion poll. For each wave $t$, let $q_t \in \mathcal{Q}$ denote the survey question asked. 
Each participant $i \in \{1,\dots,N\}$ provides a response $y_{i,t}$ to $q_t$, respecting the response space $\mathcal{Y}_{q_t}$. (e.g., multiple-choice options or free-text answers). Given a history window of length \(T-1\), we generate a synthetic response at wave \(T\) using an LLM:
\[
\hat{y}_{i,T} \;=\; f\!\big(\mathbf{x}_i,\, y_{i,1:T-1}\big).
\]
Our objective is to recover the population estimand at wave \(T\) as the finite-population mean over the unlabeled, demographic-only frame:
\[
\theta^\ast := \frac{1}{N}\sum_{i=1}^N \phi\!\left(y_{i,T}\right),
\]
where \(\phi:\mathcal{Y}\to\mathbb{R}\) maps responses to a common scale (e.g. numeric coding for Likert, scalar extraction for open-ended). Population-level estimates are, by definition, question-specific and require a single target question.

\paragraph{Synthesis.}
We use LLMs to generate survey responses, considering both inference-time adaptations (e.g., demographic or persona prompting) and training-time adaptations (e.g., domain-specific fine-tuning on survey data). At a high level, synthesis strategies divide into two categories:

Prompt-based methods rely on conditioning without parameter updates. \texttt{Demo-only} (demographic conditioning) prompts models with participant demographics $\mathbf{x}_i$ alone. \texttt{Persona-guided} extends demographic prompting by incorporating behavioral patterns from past responses $(y_{i,1:T-1})$. An auxiliary LLM analyzes each participant's past responses to generate natural language personas capturing recurring behavioral tendencies, which then condition response generation at time $T$.

Fine-tuning methods adapt model parameters using training data, typically through supervised fine-tuning (SFT) as in instruction-following setups \cite{ouyang2022training}. \texttt{Domain-FT} fine-tunes on historical responses from our target survey across time points $1$ to $T-1$, learning from (question, demographics, response) triplets via standard cross-entropy loss. \texttt{SubPOP-FT} fine-tunes on the SubPOP auxiliary dataset \cite{suh2025language}, which contains 3,229 questions from American Trends Panel with response distributions across 22 subpopulations. This method applies first-token alignment to minimize KL divergence between model logits and empirical response distributions for each (question, subgroup) pair, using external survey data rather than the target survey’s history. We include it due to its demonstrated generalization to unseen surveys and subpopulations \cite{suh2025language}.


\begin{table*}[h]
\centering
\footnotesize
\setlength{\tabcolsep}{6pt}
\renewcommand{\arraystretch}{1.15}
\begin{tabularx}{\textwidth}{@{}l *{3}{>{\centering\arraybackslash}X}@{}}
\toprule
& \textbf{NHANES (Diet)} & \textbf{ATP Q1 (Economics)} & \textbf{ATP Q2 (Politics)} \\
\midrule
Response format   & Open-ended (24h recall) & Multiple choice (6) & Multiple choice (4) \\
Participants      & 8.5k                    & 691                & 643 \\
Target            & Mean daily energy intake (kcal) & Mean Likert score & Mean Likert score \\
Target mean & 1766 kcal & 3.16 (scale 1--6) & 3.57 (scale 1--4) \\
Covariates        & 12 demo./lifestyle      & 25 demo./political  & 25 demo./political \\
Waves (\(T\)) and repeat & 2 (repeated)      & 4 (not repeated)    & 4 (not repeated) \\
\bottomrule
\end{tabularx}
\caption{\textbf{Datasets used in our evaluation.} NHANES includes two waves (\(T{=}2\)) asking the same food choice question, so responses are directly comparable across waves. ATP Q1 and Q2 are observed over four waves ($T{=}4$), with unique (non repeated) focal questions at wave $T$. We nevertheless treat ATP as a panel on covariates and prior responses: cross-wave trajectories (waves $1{:}T{-}1$) are used to construct personas and for fine-tuning.}
\label{tab:dataset_summary}
\end{table*}

\paragraph{Rectification.}
In the synthesis step, a language model \(f\) generates predictions \(\hat{y}_{i,T}\) for each participant. However, these predictions can be biased by factors such as the model’s training data or the chosen prompt \cite{bender2021dangers}. 
An alternative is to collect human responses for the same survey question to try and estimate $\theta^*$. While more reliable, such human data are costly to obtain \cite{groves2011survey}, and far less abundant than LLM responses. This trade-off motivates correction frameworks such as PPI and DSL \cite{angelopoulos2023prediction, egami2023using}, which combine cheap, plentiful model predictions with a small set of human answers. 

We therefore assume access to a small set of human responses \(\mathcal{H}=\{(\mathbf{x}_j,y_j)\}_{j=1}^n\) at wave \(T\). For each \(j\in\mathcal{H}\), we also compute a model prediction \(\hat{y}_j=f_{\pi}(\mathbf{x}_j)\) using our synthesis setup. A general correction estimator takes the form
\begin{equation}
\hat{\theta}_{\lambda}
=
\underbrace{\frac{1}{N}\sum_{i=1}^{N}\lambda\,\hat{y}_i}_{\text{synthetic mean}}
+
\underbrace{\frac{1}{n}\sum_{j=1}^{n}\bigl(y_j-\lambda\,\hat{y}_j\bigr)}_{\text{bias correction}}\, ,
\end{equation}
where \(\lambda\in[0,1]\) is a scalar (``power-tuning’’ parameter) interpolating between ignoring model predictions (\(\lambda=0\)) and using them fully (\(\lambda=1\)). This formulation corresponds to the PPI estimator \cite{angelopoulos2023prediction}, with DSL \cite{egami2023using} recovered as the special case $\lambda = 1$, which we refer to as $\text{Rec}_{\lambda = 1}$. When $\lambda$ is not specified, it is chosen from the human set $\mathcal{H}$ using the PPI++ power-tuning rule \cite{angelopoulos2023ppi++}, which minimizes the estimated variance of the estimator; we denote this as $ \text{Rec}_{\lambda_{\text{opt}}} $.

A key benefit is variance reduction. If the synthetic mean (first term) is computed on a set $\mathcal{U}$ disjoint from the human responses set $\mathcal{H}$, the variance decomposes as
\begin{equation}
\operatorname{Var}\!\bigl(\hat{\theta}_{\lambda}\bigr)
= \frac{\lambda^{2}\operatorname{Var}(\hat{y})}{N}
+ \frac{\operatorname{Var}\!\bigl(y-\lambda\,\hat{y}\bigr)}{n}\, .
\end{equation}
The first term is the variability of synthetic predictions, while the second term reflects the prediction error variance on the human responses set. According to Eq.~(2), two conditions make this estimator more effective than using human data alone: (i) access to a large set of demographics and (ii) reasonably accurate predictions. Or, formally:
\begin{equation}
\frac{\lambda^{2}\operatorname{Var}(\hat{y})}{N}
+
\frac{\operatorname{Var}\!\bigl(y-\lambda\,\hat{y}\bigr)}{n}
<
\frac{\operatorname{Var}(y)}{n}\, .
\end{equation}
The first condition is typically satisfied in survey research, since demographic covariates can be collected at scale (e.g., from census data) without requiring human responses to substantive questions. The second condition is equally important: an accurate model means the prediction error variance, $\operatorname{Var}(y - \hat{y})$, is small. As a result, the second term in Eq.~(3) becomes negligible, and the estimator's variance is dominated by the first term $\frac{\operatorname{Var}(\hat{y})}{N}$. Since this term shrinks as the synthetic sample size $N$ increases, rectification can produce significantly tighter confidence intervals than estimators that rely solely on the small set of human responses. We refer readers to Appendix~\ref{app:ppi} for more details.

\section{Experiments}
\label{sec:experiments}
Our evaluation uses two longitudinal panel surveys spanning different domains and response formats (Table~\ref{tab:dataset_summary}). Following our problem formulation, we evaluate at the question level: NHANES contributes one repeated dietary-intake item across two waves, while ATP contributes two distinct opinion items across four waves.

\paragraph{NHANES.}
The U.S. National Health and Nutrition Examination Survey 2015–2016 \cite{national2017national} is a food-consumption survey with over 16{,}000 full-day dietary recall entries from 8{,}500 participants across two waves (\(T{=}2\)). Each entry records participants’ food intake over the previous 24 hours in an open-ended format (e.g., ‘oatmeal 100g, rice 150g, banana 45g’), along with total daily energy intake. Participant metadata includes 12 demographic and lifestyle covariates (e.g., age, sex, income). This survey allows us to evaluate open-ended generation with substantial individual- and temporal-level variation. The target is mean daily energy intake (kcal) per participant, with a dataset mean of 1,766.

\paragraph{American Trends Panel (ATP).}
ATP is the Pew Research Center’s longitudinal panel for U.S. public-opinion research \cite{pew_atp}, consisting of approximately 10{,}000 randomly selected adults nationwide. We select two focal questions from waves 146–149, differing in domain (economics vs. politics) and distribution (normal vs. skewed), providing a controlled yet diverse setting to assess the robustness of the methods.

\textit{Question 1} (Economic well-being): ``Compared to your parents when they were the age you are now, do you think your own standard of living now is...'' Options: (1) Much better, (2) Somewhat better, (3) About the same, (4) Somewhat worse, (5) Much worse, (6) Not sure. We analyze 691 complete cases. The target is the mean Likert score on a 1–6 scale (dataset mean: 3.16).

\textit{Question 2} (Political opinion): ``How would you rate the job Supreme Court justices are doing in keeping their own political views out of how they decide major cases?'' Options: (1) Excellent, (2) Good, (3) Only fair, (4) Poor. We analyze 643 complete cases. The target is the mean Likert score on a 1–4 scale with a mean of 3.57.

Both ATP items include 25 demographic and political covariates (e.g., age, gender, education, race, party ID, income, region). The two questions have different answer distributions (Q1 is approximately normal, Q2 left-skewed), allowing us to test performance across distinct response patterns.

\paragraph{Evaluation setup.}
We compare four synthesis methods across multiple datasets and models, applying two post-hoc correction strategies uniformly to each. For each dataset and model, we generate synthetic responses at wave $T$ using one of the four synthesis methods described in \textsection\ref{sec:framework}. All models use a fixed sampling temperature of $\tau=0.7$, with prompts held constant (Appendix~\ref{app:prompts}). We evaluate across four language models: Qwen2.5 8B, Llama 3.1 8B, Mistral v0.3 7B, and GPT-4o mini. Rectification methods are applied on top of each synthesis strategy. At wave $T$, we draw $n_{\text{human}}=100$ participants as the gold-standard set and treat the remainder as unlabeled. As baselines, we use previous-day responses for NHANES and random responses for ATP, since its questions are not repeated across waves.

We assess performance using two complementary metrics capturing bias and variance. Bias is measured as the relative error between estimated and true population parameters:
\begin{equation}
\Delta_\% = \frac{\lvert \hat{\theta} - \theta^* \rvert}{\theta^*} \times 100,
\end{equation}
where $\hat{\theta}$ is our estimator and $\theta^*$ is the ground-truth population parameter from full human responses. Variance reduction is summarized by ESS gain,
\begin{equation}
\text{ESS}_{\text{gain}\%} = \left(\frac{\text{Var}(\hat{\theta}_{\text{human}})}{\text{Var}(\hat{\theta}_{\text{method}})} - 1\right) \times 100,
\end{equation}
so, for example, an ESS gain of 50\% means the method achieves the same precision as having 1.5$\times$ more human data. This is equivalent to getting ``more information'' out of each human response.

\begin{table*}[t]
\centering
\small
\begin{tabular}{l ccc>{\columncolor{gray!20}}c ccc>{\columncolor{gray!20}}c}
\toprule
\multirow{2}{*}{\textbf{Method}} & \multicolumn{4}{c}{\textbf{Bias (\%)$\downarrow$}} & \multicolumn{4}{c}{\textbf{ESS Gain (\%) $\uparrow$} 
} \\
\cmidrule(lr){2-5} \cmidrule(lr){6-9}
 & \textbf{NHANES} & \textbf{ATP Q1} & \textbf{ATP Q2} & \textbf{Avg.} & \textbf{NHANES} & \textbf{ATP Q1} & \textbf{ATP Q2} & \textbf{Avg.} \\
\midrule
\texttt{Baseline} & \cellcolor{red1!50}7.61 & \cellcolor{green1!50}2.30 & \cellcolor{red1!50}62.41 & 24.11 & $\dagger$ & $\dagger$ & $\dagger$ & $\dagger$ \\
\midrule
\multicolumn{9}{c}{\textit{Synthesize only}} \\
\midrule[\heavyrulewidth]
\texttt{Domain-FT | None} & \cellcolor{red1!50}7.03 & \cellcolor{red1!50}34.27 & \cellcolor{red1!50}62.69 & \textbf{34.66} & $\dagger$ & $\dagger$ & $\dagger$ & $\dagger$ \\
\texttt{SubPOP-FT | None} & \cellcolor{red1!50}180.78 & \cellcolor{red1!50}44.82 & \cellcolor{red1!50}33.08 & 86.23 & $\dagger$ & $\dagger$ & $\dagger$ & $\dagger$ \\
\texttt{Demo-only | None} & \cellcolor{red1!50}88.10 & \cellcolor{red1!50}47.53 & \cellcolor{red1!50}31.18 & 55.60 & $\dagger$ & $\dagger$ & $\dagger$ & $\dagger$ \\
\texttt{Persona-guided | None} & \cellcolor{red1!50}82.08 & \cellcolor{red1!50}50.22 & \cellcolor{red1!50}19.99 & 50.76 & $\dagger$ & $\dagger$ & $\dagger$ & $\dagger$ \\
\midrule[\heavyrulewidth]
\multicolumn{9}{c}{\textit{Synthesize + Rectify}} \\
\midrule[\heavyrulewidth]
\texttt{Domain-FT | $\text{Rec}_{\lambda = 1}$}        & \cellcolor{red1!50}9.41  & \cellcolor{red1!50}8.46  & \cellcolor{green1!50}4.11  & \textbf{7.33}  & \cellcolor{red1!50}-32.95 & \cellcolor{red1!50}-16.97 & \cellcolor{red1!50}-62.23 & -37.38 \\
\texttt{SubPOP-FT | $\text{Rec}_{\lambda = 1}$}        & \cellcolor{red1!50}37.57 & \cellcolor{red1!50}8.74  & \cellcolor{red1!50}10.98 & 19.10 & \cellcolor{red1!50}-72.91 & \cellcolor{red1!50}-13.57 & \cellcolor{red1!50}-41.69 & -42.72 \\
\texttt{Demo-only | $\text{Rec}_{\lambda = 1}$} & \cellcolor{red1!50}13.38 & \cellcolor{red1!50}16.94 & \cellcolor{green1!50}3.12  & 11.15 & \cellcolor{red1!50}-50.51 & \cellcolor{red1!50}-22.27 & \cellcolor{red1!50}-24.05 & \textbf{-32.28} \\
\texttt{Persona-guided | $\text{Rec}_{\lambda = 1}$}   & \cellcolor{red1!50}12.32 & \cellcolor{red1!50}11.10 & \cellcolor{green1!50}4.50  & 9.31  & \cellcolor{red1!50}-60.61 & \cellcolor{red1!50}-20.47 & \cellcolor{red1!50}-28.34 & -36.47 \\
\midrule    
\texttt{Domain-FT | $ \text{Rec}_{\lambda_{\text{opt}}} $}        & \cellcolor{green1!50}3.33 & \cellcolor{green1!50}1.73 & \cellcolor{green1!50}3.40 & \textbf{2.82} & \cellcolor{green1!50}2.81 & \cellcolor{green1!50}1.32 & \cellcolor{green1!50}1.01 & 1.71 \\
\texttt{SubPOP-FT | $ \text{Rec}_{\lambda_{\text{opt}}} $}        & \cellcolor{green1!50}3.23 & \cellcolor{green1!50}4.06 & \cellcolor{green1!50}3.07 & 3.45 & \cellcolor{green1!50}1.28 & \cellcolor{green1!50}1.74 & \cellcolor{green1!50}1.85 & 1.34 \\
\texttt{Demo-only | $ \text{Rec}_{\lambda_{\text{opt}}} $}        & \cellcolor{green1!50}1.75 & \cellcolor{red1!50}6.75 & \cellcolor{green1!50}4.52 & 4.34 & \cellcolor{green1!50}6.07 & \cellcolor{green1!50}4.83 & \cellcolor{green1!50}2.62 & 3.97 \\
\texttt{Persona-guided | $ \text{Rec}_{\lambda_{\text{opt}}} $}   & \cellcolor{red1!50}3.99 & \cellcolor{green1!50}4.42 & \cellcolor{red1!50}7.65 & 5.35 & \cellcolor{green1!50}14.19 & \cellcolor{green1!50}5.57 & \cellcolor{green1!50}1.01 & \textbf{6.92} \\
\bottomrule
\end{tabular}
\caption{\textbf{Bias (\%) $\downarrow$ and effective sample size (ESS) gain (\%) $\uparrow$ on three datasets}.
Top block: unrectified LLM synthesis (\textit{Synthesize only}). 
Bottom block: synthesis combined with rectification.
\texttt{X | Y} indicates synthesis method \texttt{X} with rectification method \texttt{Y}. 
All results are computed at a human sample size of $|\mathcal{H}| = n_{\text{human}}{=}100$.
Green indicates 95\% bootstrap CIs where bias includes 0 and ESS $>$ 0; averages are macro-averages across datasets. 
$\dagger$ = ESS not reported because nominal coverage (0.95) is not achieved for unrectified methods. For $ \text{Rec}_{\lambda_{\text{opt}}} $, $\lambda$ was automatically selected with average values of $\lambda \approx 0.15$ (NHANES), $0.3$ (ATP~Q1), and $0.05$ (ATP~Q2).
}
\label{tab:bias_variance}
\end{table*}

\begin{table*}[ht]
\centering
\small
\begin{tabular}{l c c c c c}
\toprule
\textbf{Subgroup} & \textbf{n} & \textbf{Bias (FT only) $\downarrow$} & \textbf{Bias (FT $|$ $ \text{Rec}_{\lambda_{\text{opt}}} $) $\downarrow$} & \textbf{Abs.\ $\Delta$ $\uparrow$} & \textbf{Rel.\ $\Delta$ (\%) $\uparrow$} \\
\midrule
sex: Female & 3608 & 4.62 & 7.32 & \cellcolor{red1!50}-2.70 & \cellcolor{red1!50}-58.55 \\
sex: Male & 3419 & 13.86 & 8.72 & \cellcolor{green1!50}5.14 & \cellcolor{green1!50}37.08 \\
race: Non-Hispanic White & 2342 & 7.93 & 4.38 & \cellcolor{green1!50}3.54 & \cellcolor{green1!50}44.71 \\
race: Non-Hispanic Black & 1525 & 5.79 & 3.45 & \cellcolor{green1!50}2.34 & \cellcolor{green1!50}40.37 \\
race: Mexican American & 1312 & 7.77 & 4.83 & \cellcolor{green1!50}2.94 & \cellcolor{green1!50}37.87 \\
household income: \$35{,}000 to \$44{,}999 & 717 & 8.34 & 4.25 & \cellcolor{green1!50}4.09 & \cellcolor{green1!50}49.07 \\
household income: \$100{,}000 and over & 1182 & 8.34 & 5.54 & \cellcolor{green1!50}2.80 & \cellcolor{green1!50}33.61 \\
\bottomrule
\end{tabular}
\caption{\textbf{Subgroup bias changes on NHANES.} 
We compare fine-tuned (FT) models before and after global rectification ($ \text{Rec}_{\lambda_{\text{opt}}} $).  
\textcolor{green1}{Green} = subgroup bias decreases (improvement); 
\textcolor{red1}{red} = subgroup bias increases (deterioration). Fixed human sample size of $n_{\text{human}}{=}100$.
Note: NHANES records ``sex'' as a self-reported variable (male/female).}
\label{tab:subgroup-bias-ft}
\end{table*}

\section{Results}
We first compare synthesis and rectification methods across datasets, focusing on population-level bias and variance (Table~\ref{tab:bias_variance}). Then, we turn to deeper analyses, examining how to best allocate human responses between fine-tuning and correction, as well as how rectification affects subgroup bias.

\begin{figure*}[ht]
    \centering
    \includegraphics[width=0.95\linewidth]{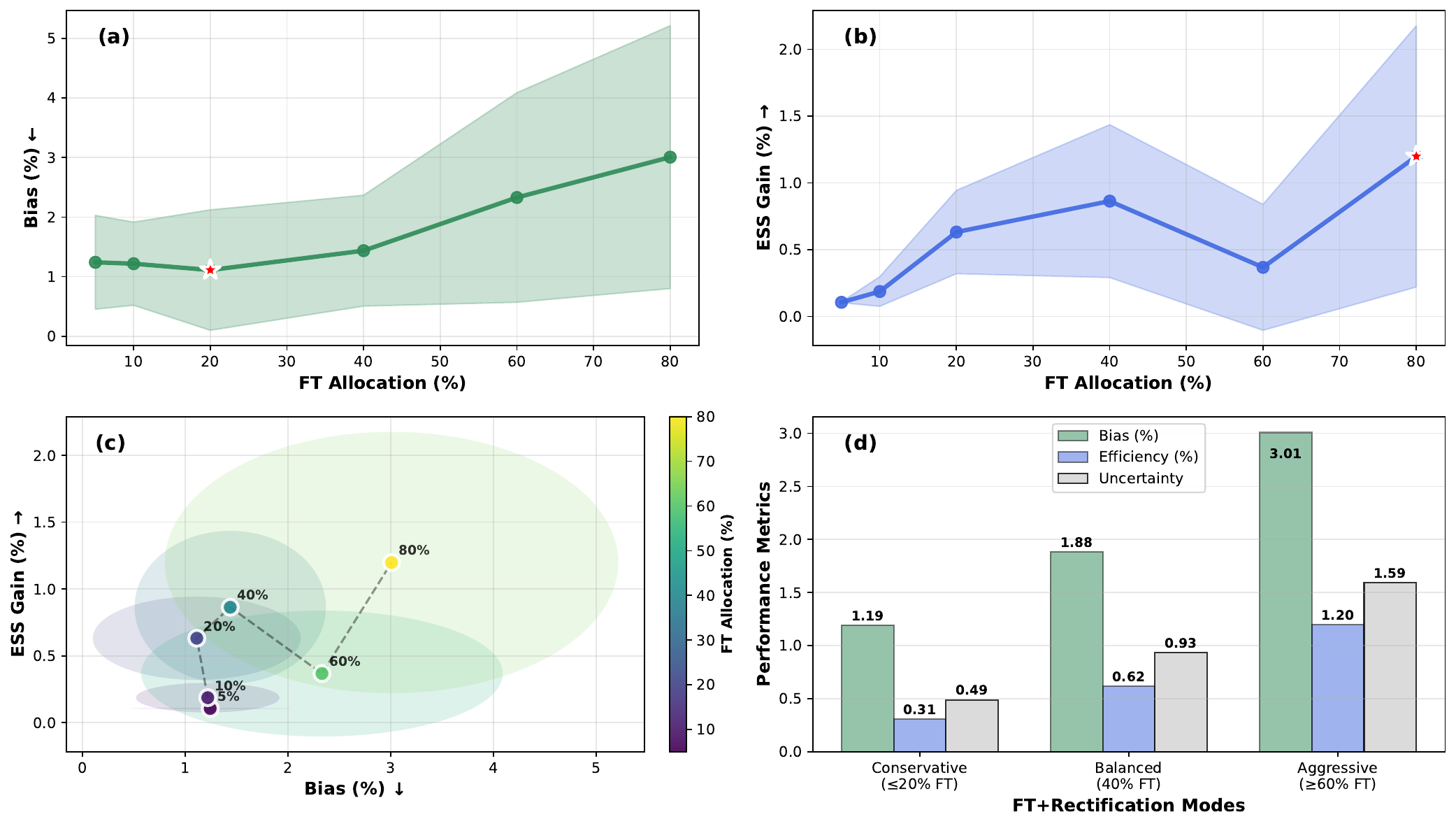}
    \caption{\textbf{Fine-tuning and rectification interaction analysis.} Results are averaged over 100 independent runs. (a) Bias vs.\ FT allocation with confidence bands; 20\% allocation minimizes bias (red star). (b) Efficiency peaks at 80\% FT allocation but with high uncertainty. (c) Pareto frontier with 95\% confidence ellipses where points toward the upper left reflect better trade-offs. (d) Comparison across allocation policies.}
    \label{fig:ft_vs_ppi}
\end{figure*}

\subsection{Bias and efficiency across methods}
\label{sec:results1}

\uline{Unrectified synthesis is biased, whereas rectification consistently fixes it.} 
Pure LLM synthesis shows large and inconsistent bias (top block). Averaged across datasets, the baseline has 24.11\% bias, while \texttt{Domain-FT}, \texttt{Persona-guided}, \texttt{Demo-only}, and \texttt{SubPOP-FT} have 34.66\%, 50.76\%, 55.60\%, and 86.23\%, respectively. Per-dataset behavior is heterogeneous: e.g., on ATP~Q2 (skewed/heavy-tailed answer distribution), \texttt{Persona-guided} reduces the unrectified bias to 19.99\% vs.\ a 62.41\% baseline, but performs poorly elsewhere. This variability in bias across methods and datasets confirms that synthesis-only methods cannot be trusted to make valid claims about population preferences or behaviors. Applying rectification collapses bias to single digits (bottom block). $\text{Rec}_{\lambda = 1}$ achieves some reduction but leaves significant residual bias, making it less reliable than $ \text{Rec}_{\lambda_{\text{opt}}} $. By contrast, $ \text{Rec}_{\lambda_{\text{opt}}} $ consistently drives bias even lower: the lowest average bias is achieved by \texttt{Domain-FT | $ \text{Rec}_{\lambda_{\text{opt}}} $} (2.82\%), followed closely by \texttt{SubPOP-FT | $ \text{Rec}_{\lambda_{\text{opt}}} $} (3.45\%), both training-based methods, while all approaches under $ \text{Rec}_{\lambda_{\text{opt}}} $ on average remain below 5.5\% bias. Importantly, training-based methods with $ \text{Rec}_{\lambda_{\text{opt}}} $ reduce bias to statistically insignificant levels across all datasets, unlike other combinations.

\uline{$ \text{Rec}_{\lambda_{\text{opt}}} $ rectification enables positive ESS gains.}
ESS is meaningful only when confidence intervals achieve nominal coverage, i.e., when they contain the true population value at the expected frequency. All unrectified methods fail this criterion. However, when combined with $ \text{Rec}_{\lambda_{\text{opt}}} $, every method achieves positive ESS gains, confirming a significant reduction in variance relative to the human-only estimator. \texttt{Persona-guided} + $ \text{Rec}_{\lambda_{\text{opt}}} $ attains the highest average ESS gain (6.92\%) and peaks at 14.19\% on NHANES, equivalent to a $\approx$14\% increase in human sample size without additional data collection. In contrast, $\text{Rec}_{\lambda = 1}$ consistently produces negative ESS gains. As expected, the most biased methods deliver the largest ESS gains, a direct manifestation of the bias–variance trade-off. Based on these findings, we focus on $ \text{Rec}_{\lambda_{\text{opt}}} $ in all subsequent analyses.

\uline{ESS gains shrink as the number of human responses grow.}
We next vary the number of human responses used in rectification $ \text{Rec}_{\lambda_{\text{opt}}} $, $n_{\text{human}}\!\in\!\{50,100,150,200\}$, and track ESS (Fig.~\ref{fig:ess_gain}, Appendix~\ref{app:ess-gain}). Similarly, \texttt{Persona-guided} maintains the largest ESS gains across settings, while all methods show an expected decline in ESS gain as $n_{\text{human}}$ increases (the human-only estimator improves). In practice, correction is most beneficial when $n_{\text{human}}$ is limited and the remaining pool $N$ is large. A reasonable stopping rule in terms of variance is to discontinue using synthetic responses once Eq.~(3) is not satisfied.

\subsection{Subgroup effects and allocation strategies }
\label{sec:results2}
\uline{Rectification reduces bias per subgroup.}
A key concern is whether rectification reduces error consistently across sub-populations, rather than only at the population level. To test this, we evaluate subgroup error on NHANES before and after applying global $ \text{Rec}_{\lambda_{\text{opt}}} $ rectification to a fine-tuned model (Table~\ref{tab:subgroup-bias-ft}). Specifically, we re-center each response around the estimate~$\theta_{\text{$ \text{Rec}_{\lambda_{\text{opt}}} $}}$ and compare subgroup-level bias. Most subgroups (6 of 7) show reductions (e.g., Mexican-Americans: --37\%; income \$35--45k: --49\%), though bias increases for female respondents (+58\%). Thus, while population-level re-centering renders such heterogeneity expected, careful validation is required to ensure no subgroup suffers increased bias. Nevertheless, the overall trend suggests that rectification mitigates performance disparities across groups.

\uline{Rectification for bias, fine-tuning for efficiency.}
\label{sec:ft_vs_ppi}
Since training-based methods achieve the best rectified performance, a natural question arises: with a limited human set, should responses be allocated to fine-tuning the synthesis model or reserved for post-hoc correction\footnote{PPI requires held-out data; reusing fine-tuning data violates its guarantees~\cite{angelopoulos2023prediction}.}? To study this trade-off, we fix a budget of 1,000 human responses and evaluate allocations between \texttt{Domain-FT} and $ \text{Rec}_{\lambda_{\text{opt}}} $ in proportions of 10–90, 20–80, 40–60, 60–40, and 80–20. Experiments are run on NHANES, as it offers a larger sample size than ATP \footnote{This analysis uses a different data split than Table~\ref{tab:bias_variance}.}. 

Figure~\ref{fig:ft_vs_ppi} shows that bias is lowest when 20\% of responses are allocated to fine-tuning and the remainder to correction (panel a). Larger allocations (40--80\%) yield greater efficiency gains (panel b) but at the cost of higher bias and uncertainty. The Pareto frontier (panel c) summarizes this trade-off: points toward the upper left represent more favorable combinations. To illustrate, we group strategies into three regimes (panel d): \emph{Conservative} ($\leq$200 responses for FT), \emph{Balanced} ($\approx$400 for FT), and \emph{Aggressive} ($\geq$600 for FT). These regimes highlight how different allocation rules trace distinct positions along the frontier, offering interpretable levers for balancing bias and efficiency.



\section{Discussion}
\label{sec:discussion}

Across three longitudinal surveys, we find that all adaptation strategies reduce LLM bias once rectified, with \texttt{Domain-FT} achieving the lowest error ($<$3\% on average). At the same time, every synthesis method yields significant ESS gains after correction, showing clear improvements over a human-only estimator. 

Augmenting surveys with LLM responses is therefore not a universal solution but a set of context-dependent choices guided by the analyst’s goals. Accordingly, we present our recommendations in Table~\ref{tab:guidelines} as practical guidelines. These guidelines are supported by experiments in nutrition, economics, and politics within the U.S., where LLMs show reliable factual grounding. While we expect the main patterns to hold broadly, generalizing to other domains requires awareness of potential differences in performance.


Overall, we show that rectification improves efficiency when a large demographic frame is available and model predictions are reasonably accurate (Eq.~3). Power-tuning further enhances efficiency by avoiding negative ESS outcomes, unlike $\text{Rec}_{\lambda = 1}$. Subgroup analysis suggests that it reduces disparities, but warrants further validation.

Moving forward, promising directions include developing subgroup-aware correction methods, extending evaluations to multilingual and cross-cultural contexts, and testing live deployments within survey infrastructure. Beyond surveys, our findings highlight how LLM predictions and limited human data can be combined for valid inference in other settings where labeled data are scarce.

\begin{table}[ht!]
\centering
\small
\setlength{\tabcolsep}{4pt}
\colorbox{softblue!90}{%
  \begin{tabular}{@{}>{\centering\arraybackslash}m{0.25\linewidth} >{\centering\arraybackslash}m{0.70\linewidth}@{}}
  \rowcolor{softblue!60}
  \textbf{Objective} & \textbf{Practical recommendations} \\
  \midrule[0.5pt]
  \textbf{A. Minimizing the error of the population estimate} &
  \begin{minipage}[t]{\linewidth}
  \vspace{1pt}
  \begin{itemize}[leftmargin=1.2em, itemsep=0pt, topsep=0pt, parsep=0pt, partopsep=0pt, label=\raisebox{0.2ex}{\scriptsize$\bullet$}]
    \item Reserve the majority of the labeling budget for post-hoc correction.
    \item Synthesize with \texttt{Domain-FT} (or \texttt{SubPOP-FT} if target histories are unavailable).
  \end{itemize}
  \strut\vspace{0.5pt}
  \end{minipage} \\
  \hdashline
  \textbf{B. Minimizing the number of needed human responses} &
  \begin{minipage}[t]{\linewidth}
  \vspace{1pt}
  \begin{itemize}[leftmargin=1.2em, itemsep=0pt, topsep=0pt, parsep=0pt, partopsep=0pt, label=\raisebox{0.2ex}{\scriptsize$\bullet$}]
    \item Similarly allocate the majority share to correction.
    \item Afterwards, generate responses with \texttt{Persona-guided} prompting for the largest ESS gains.
  \end{itemize}
  \strut\vspace{0.5pt}
  \end{minipage} \\
  \hdashline
  \textbf{C. Inference in very low data or compute regimes} &
  \begin{minipage}[t]{\linewidth}
  \vspace{1pt}
  \begin{itemize}[leftmargin=1.2em, itemsep=0pt, topsep=0pt, parsep=0pt, partopsep=0pt, label=\raisebox{0.2ex}{\scriptsize$\bullet$}]
    \item When neither historical data nor fine-tuning compute are available, apply \texttt{Demo-only} synthesis plus correction.
    \item Bias collapses below 5\% and ESS improves modestly. 
  \end{itemize}
  \strut\vspace{0.5pt}
  \end{minipage} \\
  \hdashline
  \textbf{D. Having the best synthesis model for follow-up} &
  \begin{minipage}[t]{\linewidth}
  \vspace{1pt}
  \begin{itemize}[leftmargin=1.2em, itemsep=0pt, topsep=0pt, parsep=0pt, partopsep=0pt, label=\raisebox{0.2ex}{\scriptsize$\bullet$}]
    \item If the end-goal is a single best-performing synthesis model (e.g., for follow-up questions or downstream quantities beyond the corrected estimate), skip prompting and \texttt{fine-tune} directly on available responses. 
  \end{itemize}
  \strut\vspace{0.5pt}
  \end{minipage} \\
  \bottomrule[0.9pt]
  \end{tabular}%
}
\caption{\textbf{Evidence-based guidelines for LLM-assisted survey simulation.}}
\label{tab:guidelines}
\end{table}






\section{Conclusion}

We presented the first head-to-head evaluation of training-time, inference-time, and post-hoc adaptations for LLM-assisted survey simulation across three longitudinal datasets. Our results show that uncorrected LLM synthesis is consistently biased, but rectification with a small set of human responses reduces bias to below 5\% while achieving positive ESS gains. Training-based methods paired with rectification ($ \lambda_{\text{opt}} $) provide the most reliable estimates, while inference-time prompting strategies deliver greater ESS gains at the cost of higher bias. In our $N{=}1000$ experiments, allocating a majority of responses (60--80\%) to correction gave the best trade-off between bias and efficiency, though the precise percentages will vary with data and budget.

\section*{Limitations}

We identify key limitations of our work. First, our evaluation is restricted to the U.S.\ context, drawing on NHANES and ATP data with all interactions conducted in English. Performance is likely to vary across languages, cultural norms, and survey methodologies \cite{shi2024culturebank,russo2023spillover, ziegenfuss2021impact}. Extending benchmarks to multilingual and non-Western contexts is therefore essential before drawing global conclusions.  

Second, our experiments rely on simple non-adaptive correction methods. Although accessible due to their simplicity and theoretical guarantees, alternative methods such as confidence-driven inference (CDI) \cite{gligoric2024can} may offer stronger performance in practice. In particular, adaptive procedures that re-weight based on model confidence could deliver larger efficiency gains, especially when prediction reliability varies across subgroups. However, adaptive approaches during data collection (rather than post hoc) are less accessible to practitioners and require careful validation of the sampling rule. Future work should examine how practitioners balance data-collection simplicity against potential efficiency gains.

Third, rectification is effective for population-level estimation but does not solve the harder problem of individual-level simulation. Accurately reproducing a single respondent’s answers requires capturing idiosyncratic confounders and latent traits \cite{shaikh2025creating,belyaeva2023multimodal}, which current methods struggle to represent. We explicitly show the difficulty of this problem through the results in Appendix \ref{app:individual-results}. Progress here will likely require richer behavioral models and new data sources. Moreover, current rectification methods operate as numerical adjustments, while the actual responses themselves remain biased (i.e., we do not directly correct the LLM outputs). Future work could explore approaches that jointly improve both the statistical estimates and the generated responses.


\section*{Ethical considerations}



Survey data often include sensitive personal information. If such data are processed by large language models hosted by commercial providers, questions of privacy and consent become paramount~\cite{kalluri2025computer}. Participants should retain meaningful control over their responses, and safeguards are needed to prevent concentration of power among technology companies~\cite{vincent2021data,vincent2023chatgpt}.

Replacing or reducing human survey participation has economic implications and potential income displacements. Surveys currently provide paid opportunities for respondents~\cite{gray2019ghost}, and widespread substitution with LLM-generated data could displace this source of income~\cite{shao2025future,tiwari2023impact}. Any deployment of methods as described in our guidelines must weigh efficiency gains against potential harms to individuals who rely on survey participation.

Large-scale use of LLMs has broader societal costs, including environmental impacts from training and inference~\cite{wu2022sustainable,zhong2024impact}. The studied approach offers a partial mitigation: by rectifying predictions from existing models rather than training new ones from scratch, we reduce the need for additional large-scale model development~\cite{lacoste2019quantifying}. In this sense, our evaluation demonstrates how methodological innovation can align with more sustainable AI practices.

Lastly, subgroup bias amplification is an important ethical consideration. If the labeled data is sparse or unrepresentative, rectification can inadvertently amplify subgroup-level errors, correcting toward majority patterns while leaving minority responses systematically misestimated. This risk is particularly salient for survey applications where small subpopulations are of substantive interest. Future work should examine subgroup-aware, stratified, or adaptive rectification strategies that explicitly mitigate disparities~\cite{fogliato2024framework, russo2020control}.

\bibliography{custom}
\newpage
\appendix

\section{Supplementary results and experiments}
\label{app:supp-results}

Both datasets used in our study are publicly available. The U.S. National Health and Nutrition Examination Survey (NHANES) is produced by the National Center for Health Statistics and is in the public domain. The American Trends Panel (ATP) is released by the Pew Research Center for scholarly use under its data use terms. No special licenses or permissions were required for access or use of these datasets in our work.
\subsection{Individual-level simulation results}
\label{app:individual-results}
This analysis supports the discussion within the limitations section regarding the difficulty of capturing idiosyncratic behaviors. We report Mean Absolute Error (MAE) between the ground-truth scalar target $y_i=\phi(Y_{i,T})$ and the synthetic value $\hat{y}_i=\phi(\hat{Y}_{i,T})$ for each individual $i$ in NHANES.
\begin{figure*}[htbp]
\centering
\includegraphics[width=0.7\textwidth]{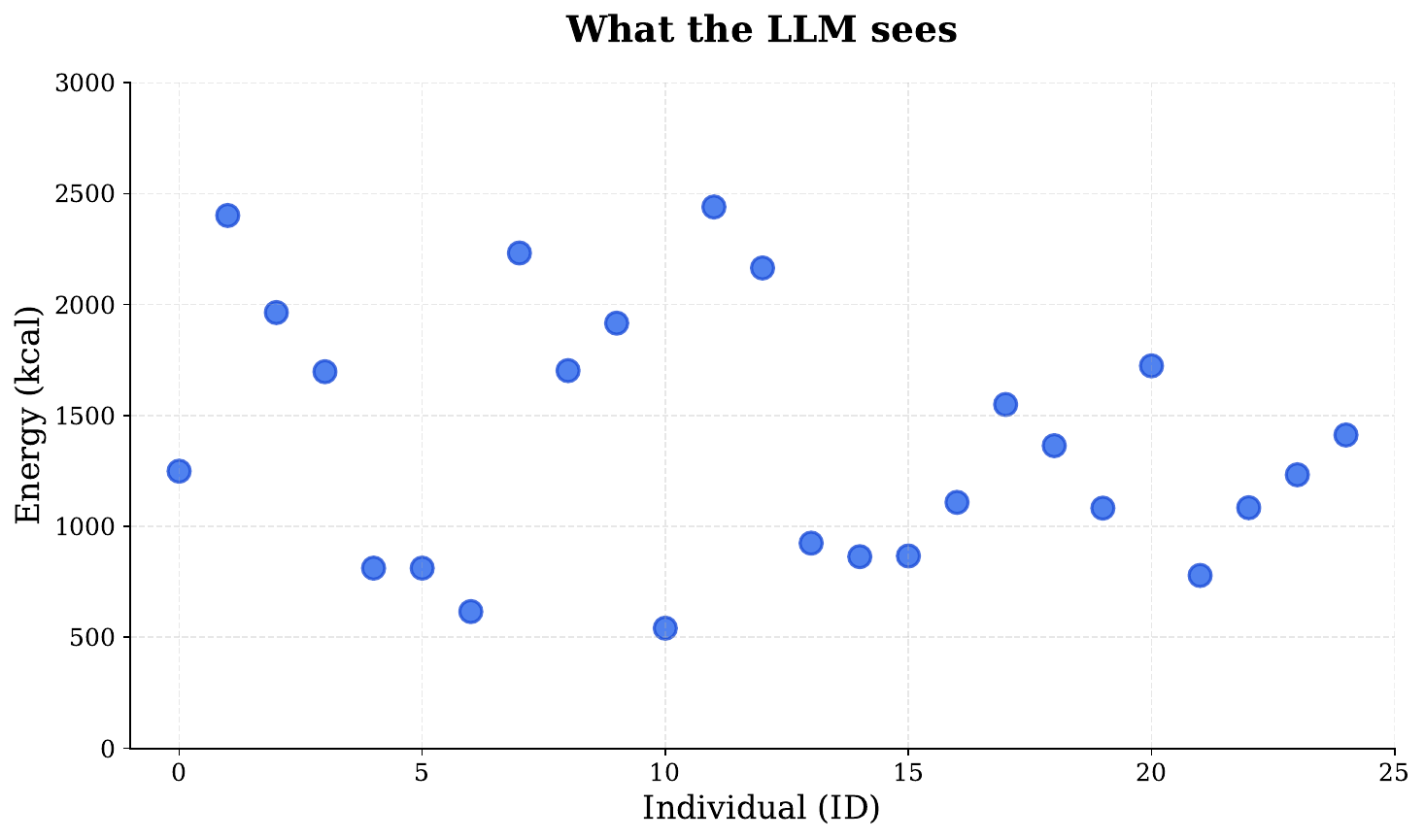}
\caption{Individual variation in energy intake within identical demographic groups. 
To the LLM, each point represents an individual with the same observable characteristics 
(age 66–70, female, 60–70kg, Non-Hispanic White, no special diet), 
yet their actual energy consumption varies greatly from 500 to 2,500 kcal.}
\label{fig:individual-challenge}
\end{figure*}


These results show that while LLMs can be effectively calibrated to produce unbiased population estimates, accurately predicting an individual's response remains fundamentally difficult. From the LLM's perspective, individuals sharing identical demographic profiles appear indistinguishable, as illustrated in Figure~\ref{fig:individual-challenge}. The model has no mechanism to differentiate between a 67-year-old white female who consumes 800 kcal versus another with identical observable characteristics who consumes 2,200 kcal.

Several factors contribute to this issue. First, \textit{identical observable features} mean that LLMs only observe coarse demographic categories, missing subtle but crucial individual differences in metabolism, food preferences, cooking skills, or economic circumstances. Second, \textit{natural daily variation} ensures that even the same individual has substantial day-to-day fluctuations based on work schedule, social events, mood, stress levels, and purely stochastic factors. Finally, \textit{unobserved determinants} such as genetics, medication effects, micronutrient status, food allergies, and personal dietary history remain completely hidden from the model yet strongly influence behavior.

\begin{table}[htbp]
\centering
\begin{tabular}{lc}
\toprule
\textbf{Synthesis method} & \textbf{MAE (kcal) $\downarrow$} \\
\midrule
\texttt{Demo-only}       & 1257.2 \\
\texttt{Persona-guided}  & 1417.5 \\
\texttt{Domain-FT}       & 815.0 \\
\texttt{SubPOP-FT}       & 920.1 \\
\bottomrule
\end{tabular}
\caption{Individual-level absolute error (MAE) for daily energy intake (kcal) on NHANES. 
The high individual-level errors highlight the difficulty of this setting.}
\label{tab:individual-results-mae}
\end{table}

\subsection{ESS gains with increasing $n_{\text{human}}$}
\label{app:ess-gain}

\begin{figure*}[ht]
    \centering
    \includegraphics[width=1.0\linewidth]{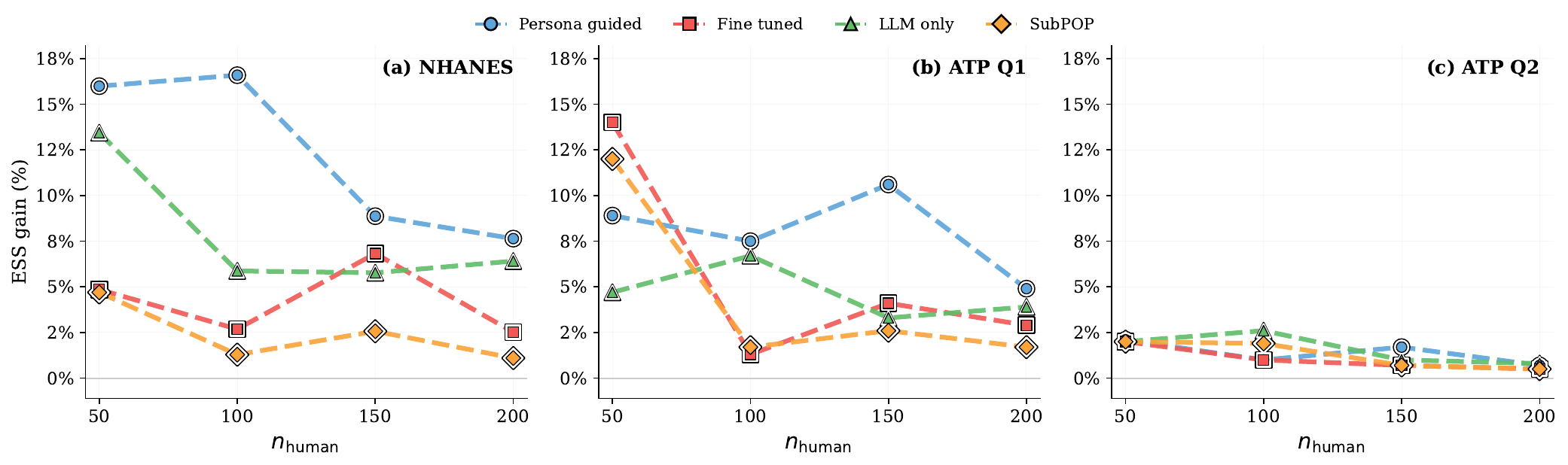}
    \caption{ESS gain under $ \text{Rec}_{\lambda_{\text{opt}}} $ across labeled-sample sizes $n_{\text{human}}\in\{50,100,150,200\}$.}
    \label{fig:ess_gain}
\end{figure*}

Figure~\ref{fig:ess_gain} reports ESS gains under $ \text{Rec}_{\lambda_{\text{opt}}} $ as the number of labeled participants increases. We observe the same trend across all three datasets: gains shrink as $n_{\text{human}}$ grows, since the human-only estimator improves with more labels.

\subsection{Ablation studies (NHANES)}
\label{app:ablations}

\noindent \textbf{Context ablation.}
We ablate the demographic information provided to the model at inference time by incrementally expanding the prompt (e.g., the demographic profile of the person). Each group adds a new set of attributes to the prompt. Group 1 uses only age and sex. Group 2 adds anthropometric variables (height, weight, BMI). Group 3 includes dietary preference (e.g., vegan, vegetarian). Group 4 incorporates race/ethnicity, and Group 5 adds citizenship status. Adding more context about the participant increases performance. The results are summarized in Table~\ref{tab:context-ablation}.

\begin{table*}[ht]
\centering
\begin{tabular}{lccc}
\toprule
\textbf{Conditioning level} & $\Delta$ from truth (\%) & CI width (\%) & CI coverage $\uparrow$ \\
\midrule
Basic: age, sex           & 13.44 & 13.42 & 0.22 \\
+ anthropometrics         & 12.34 & 15.79 & 0.34 \\
+ dietary preference      & 10.26 & 15.47 & 0.45 \\
+ ethnicity               & 11.51 & 15.57 & 0.53 \\
+ citizenship             &  8.96 & 14.55 & 0.60 \\
\bottomrule
\end{tabular}
\caption{Effect of conditioning level on estimation.}
\label{tab:context-ablation}
\end{table*}

\noindent \textbf{Prompting strategy ablation.}
We compare four prompting strategies that vary along two axes: (1) single-turn vs. multi-turn prompting, and (2) unconditioned vs. conditioned inputs. In single-turn prompting, the model is asked to recall the entire day's intake in one pass. Multi-turn prompting follows a structured format modeled after real dietary surveys (e.g., the Automated Multiple-Pass Method), where the model is guided through multiple passes and explicitly asked whether anything was forgotten. Conditioning improves accuracy and coverage. Interestingly, multi-turn prompting consistently leads to higher nutrient estimates, which may reflect memory probing (i.e., the model ``recalling'' additional foods) or acquiescence bias\footnote{Acquiescence bias is the tendency to provide affirmative responses regardless of content. When repeatedly asked ``Did you forget anything?'', LLMs may generate additional foods due to the prompting structure rather than genuine recall.}. 

\begin{table*}[ht]
\centering
\begin{tabular}{lccc}
\toprule
\textbf{Prompt format} & $\Delta$ from truth (\%) & CI width (\%) & CI coverage $\uparrow$ \\
\midrule
Single-turn, no profile   & 19.84 &  8.37 & 0.16 \\
Multi-turn, no profile    & 34.67 &  7.85 & 0.14 \\
Single-turn, full profile & 10.00 & 14.77 & 0.60 \\
Multi-turn, full profile  & 25.31 & 13.71 & 0.27 \\
\bottomrule
\end{tabular}
\caption{Comparison of prompting strategies.}
\label{tab:prompting-ablation}
\end{table*}

\subsection{Results by base model}
Table~\ref{tab:appendix_models} reports bias and ESS gains separately for each base model and synthesis–rectification combination. Overall, the patterns are consistent with our main findings: unrectified synthesis remains biased and fails to achieve valid coverage (hence ESS is not reported), while applying rectification collapses bias and yields positive ESS gains in most cases.

\begin{table*}[t]
\centering
\small
\setlength{\tabcolsep}{5pt}
\begin{tabular}{l cc c cc c}
\toprule
\multirow{2}{*}{\textbf{Model: \texttt{Synthesis | Rectify}}} & \multicolumn{3}{c}{\textbf{Bias (\%) $\downarrow$}} & \multicolumn{3}{c}{\textbf{ESS Gain (\%) $\uparrow$}} \\
\cmidrule(lr){2-4} \cmidrule(lr){5-7}
 & \textbf{NHANES} & \textbf{ATP Q1} & \textbf{ATP Q2}
 & \textbf{NHANES} & \textbf{ATP Q1} & \textbf{ATP Q2} \\
\midrule
\multicolumn{7}{l}{\textit{Unrectified synthesis (LLM-only)}} \\
\midrule
Baseline: \texttt{— | None}                         & 7.51 & 2.16 & 59.90 & $\dagger$ & $\dagger$ & $\dagger$ \\
GPT-4o-mini: \texttt{Persona-guided | None}         & 16.77 & 52.74 & 27.53 & $\dagger$ & $\dagger$ & $\dagger$ \\
GPT-4o-mini: \texttt{Demo-only | None}       & 18.11 & 32.78 & 36.77 & $\dagger$ & $\dagger$ & $\dagger$ \\
Llama: \texttt{Persona-guided | None}               & 186.24 & 63.04 & 8.66  & $\dagger$ & $\dagger$ & $\dagger$ \\
Llama: \texttt{Domain-FT | None}                    & 9.22 & 33.09 & 73.51 & $\dagger$ & $\dagger$ & $\dagger$ \\
Llama: \texttt{Demo-only | None}             & 213.77 & 83.24 & 12.67 & $\dagger$ & $\dagger$ & $\dagger$ \\
Llama: \texttt{SubPOP-FT | None}                    & 165.42 & 55.56 & 16.82 & $\dagger$ & $\dagger$ & $\dagger$ \\
Mistral: \texttt{Persona-guided | None}             & 90.15 & 52.74 & 27.53 & $\dagger$ & $\dagger$ & $\dagger$ \\
Mistral: \texttt{Domain-FT | None}                  & 2.74 & 50.57 & 71.14 & $\dagger$ & $\dagger$ & $\dagger$ \\
Mistral: \texttt{Demo-only | None}           & 108.27 & 61.00 & 44.93 & $\dagger$ & $\dagger$ & $\dagger$ \\
Mistral: \texttt{SubPOP-FT | None}                  & 359.05 & 49.66 & 47.83 & $\dagger$ & $\dagger$ & $\dagger$ \\
Qwen: \texttt{Persona-guided | None}                & 35.12 & 37.60 & 32.57 & $\dagger$ & $\dagger$ & $\dagger$ \\
Qwen: \texttt{Domain-FT | None}                     & 9.21 & 20.67 & 46.31 & $\dagger$ & $\dagger$ & $\dagger$ \\
Qwen: \texttt{Demo-only | None}              & 11.93 & 14.49 & 36.77 & $\dagger$ & $\dagger$ & $\dagger$ \\
Qwen: \texttt{SubPOP-FT | None}                     & 17.76 & 29.12 & 36.77 & $\dagger$ & $\dagger$ & $\dagger$ \\
\midrule[\heavyrulewidth]
\multicolumn{7}{c}{\textit{Synthesize, then Rectify}} \\
\midrule[\heavyrulewidth]
GPT-4o-mini: \texttt{Persona-guided | $ \text{Rec}_{\lambda_{\text{opt}}} $}          & 0.48 & 6.09 & 2.63 & 22.4 & 6.6 & 1.0 \\
GPT-4o-mini: \texttt{Persona-guided |  $\text{Rec}_{\lambda = 1}$}          & 7.59 & 6.85 & 3.48 & -41.0 & -19.1 & -25.1 \\
GPT-4o-mini: \texttt{Demo-only | $ \text{Rec}_{\lambda_{\text{opt}}} $}        & 4.50 & 7.49 & 3.69 & 4.8  & 7.6 & 1.8 \\
GPT-4o-mini: \texttt{Demo-only |  $\text{Rec}_{\lambda = 1}$}        & 5.18 & 8.83 & 2.43 & -30.9 & -24.9 & -26.0 \\
Llama: \texttt{Persona-guided | $ \text{Rec}_{\lambda_{\text{opt}}} $}                & 10.03 & 6.09 & 1.16 & 15.2 & 4.6 & 1.0 \\
Llama: \texttt{Persona-guided |  $\text{Rec}_{\lambda = 1}$}                & 35.25 & 0.81 & 7.78 & -84.6 & -29.5 & -31.0 \\
Llama: \texttt{Domain-FT | $ \text{Rec}_{\lambda_{\text{opt}}} $}                     & 4.01 & 15.01 & 6.27 & 5.4 & 1.0 & 1.0 \\
Llama: \texttt{Domain-FT |  $\text{Rec}_{\lambda = 1}$}                     & 12.67 & 14.80 & 3.63 & -19.9 & -30.4 & -60.1 \\
Llama: \texttt{Demo-only | $ \text{Rec}_{\lambda_{\text{opt}}} $}              & 4.48 & 12.07 & 2.06 & 9.5 & 8.6 & 5.1 \\
Llama: \texttt{Demo-only |  $\text{Rec}_{\lambda = 1}$}              & 20.87 & 14.40 & 0.03 & -87.8 & 7.1 & -26.0 \\
Llama: \texttt{SubPOP-FT | $ \text{Rec}_{\lambda_{\text{opt}}} $}                     & 2.43 & 9.98 & 1.10 & 1.0 & 1.2 & 1.0 \\
Llama: \texttt{SubPOP-FT |  $\text{Rec}_{\lambda = 1}$}                     & 20.09 & 12.29 & 0.38 & -85.8 & -31.3 & -64.9 \\
Mistral: \texttt{Persona-guided | $ \text{Rec}_{\lambda_{\text{opt}}} $}              & 3.08 & 5.15 & 3.69 & 25.7 & 6.6 & 1.0 \\
Mistral: \texttt{Persona-guided |  $\text{Rec}_{\lambda = 1}$}              & 3.39 & 6.85 & 2.24 & -68.0 & -12.0 & -25.1 \\
Mistral: \texttt{Domain-FT | $ \text{Rec}_{\lambda_{\text{opt}}} $}                   & 9.97 & 1.23 & 15.07 & 1.0 & 1.0 & 1.0 \\
Mistral: \texttt{Domain-FT |  $\text{Rec}_{\lambda = 1}$}                   & 13.26 & 6.30 & 4.12 & -62.9 & -22.8 & -61.3 \\
Mistral: \texttt{Demo-only | $ \text{Rec}_{\lambda_{\text{opt}}} $}            & 11.99 & 23.38 & 7.69 & 5.4 & 5.9 & 1.8 \\
Mistral: \texttt{Demo-only | $\text{Rec}_{\lambda = 1}$}            & 20.80 & 25.08 & 6.91 & -57.8 & -6.7 & -47.8 \\
Mistral: \texttt{SubPOP-FT | $ \text{Rec}_{\lambda_{\text{opt}}} $}                   & 7.31 & 5.77 & 10.02 & 1.0 & 2.0 & 3.5 \\
Mistral: \texttt{SubPOP-FT |  $\text{Rec}_{\lambda = 1}$}                   & 90.98 & 13.74 & 26.34 & -98.5 & 1.0 & -60.1 \\
Qwen: \texttt{Persona-guided | $ \text{Rec}_{\lambda_{\text{opt}}} $}                 & 3.84 & 13.39 & 2.63 & 3.1 & 11.2 & 1.0 \\
Qwen: \texttt{Persona-guided |  $\text{Rec}_{\lambda = 1}$}                 & 3.07 & 25.63 & 3.48 & -44.4 & -19.1 & -13.8 \\
Qwen: \texttt{Domain-FT | $ \text{Rec}_{\lambda_{\text{opt}}} $}                      & 4.75 & 3.40 & 1.16 & 1.6 & 2.0 & 1.0 \\
Qwen: \texttt{Domain-FT |  $\text{Rec}_{\lambda = 1}$}                      & 2.30 & 4.27 & 4.58 & -34.1 & -25.3 & -54.1 \\
Qwen: \texttt{Demo-only | $ \text{Rec}_{\lambda_{\text{opt}}} $}               & 9.83 & 20.44 & 3.69 & 3.8 & 4.7 & 1.0 \\
Qwen: \texttt{Demo-only |  $\text{Rec}_{\lambda = 1}$}               & 6.69 & 19.44 & 2.43 & -35.9 & -35.3 & 0.9 \\
Qwen: \texttt{SubPOP-FT | $ \text{Rec}_{\lambda_{\text{opt}}} $}                      & 1.35 & 1.55 & 6.22 & 1.8 & 2.8 & 1.0 \\
Qwen: \texttt{SubPOP-FT |  $\text{Rec}_{\lambda = 1}$}                      & 1.65 & 0.19 & 6.22 & -38.9 & -18.0 & 1.0 \\
\bottomrule
\end{tabular}
\caption{\textbf{Per-model results across datasets.} Bias (\%) $\downarrow$ and ESS gain (\%) $\uparrow$ (ESS@100) on NHANES, ATP~Q1, and ATP~Q2. Notation \texttt{X | Y} indicates synthesis method \texttt{X} combined with rectification method \texttt{Y}; \texttt{None} denotes unrectified synthesis. $\dagger$ = ESS not reported because nominal coverage (0.95) is not achieved. Prompting temperature fixed at 0.7 for all generations.}
\label{tab:appendix_models}
\end{table*}
\clearpage
\clearpage
\section{Methods}

\subsection{Prediction-Powered Inference (PPI)}
\label{app:ppi}
\newcommand{\E}{\mathbb{E}}
Prediction-Powered Inference (PPI) \citep{angelopoulos2023prediction} provides valid confidence intervals for statistical estimands by combining a small labeled dataset with predictions from a machine-learning system on a large unlabeled dataset. Let $\{(X_i,Y_i)\}_{i=1}^n$ denote the labeled data and $\{\tilde{X}_i\}_{i=1}^N$ denote unlabeled covariates with $N \gg n$, both drawn i.i.d.\ from the same distribution. The predictor $f$ is trained independently of both datasets.
PPI applies to estimands $\theta^*$ defined as solutions to convex optimization problems of the form:
\begin{equation*}
\theta^* = \arg\min_{\theta\in\Theta} \E[\ell_\theta(X,Y)],
\end{equation*}
where $\ell_\theta$ is a convex loss function. Under mild regularity conditions, $\theta^*$ can be characterized by the estimating equation $\E[g_{\theta^*}(X,Y)] = 0$, where $g_\theta = \nabla_\theta \ell_\theta$ is a subgradient of the loss.
PPI constructs an estimator by combining two components. The \textit{imputed gradient} uses unlabeled data and model predictions:
\begin{equation*}
\hat{g}_\theta^f = \frac{1}{N}\sum_{i=1}^N g_\theta\big(\tilde{X}_i, f(\tilde{X}_i)\big).
\end{equation*}
The \textit{rectifier} uses labeled data to correct for prediction bias:
\begin{equation*}
\widehat{\Delta}_\theta = \frac{1}{n}\sum_{i=1}^n \Big( g_\theta(X_i,Y_i) - g_\theta(X_i,f(X_i)) \Big).
\end{equation*}
The PPI estimator $\hat{\theta}_{\mathrm{PPI}}$ solves:
\begin{equation*}
\hat{g}_\theta^f + \widehat{\Delta}_\theta = 0.
\end{equation*}
This estimator is unbiased by construction:
\begin{align*}
\E[\hat{g}_{\theta^*}^f + \widehat{\Delta}_{\theta^*}]
  &= \E[g_{\theta^*}(X,f(X))]  \notag\\
  &\quad +\, \E[g_{\theta^*}(X,Y) - g_{\theta^*}(X,f(X))] \notag\\
  &= \E[g_{\theta^*}(X,Y)] = 0.
\end{align*}

\paragraph{Power tuning.}
PPI can be extended to include a power tuning parameter $\lambda \in [0,1]$ that controls the relative weight given to predictions versus labeled data. The rectifier can be scaled as $\lambda \widehat{\Delta}_\theta$, yielding the modified estimating equation $\hat{g}_\theta^f + \lambda \widehat{\Delta}_\theta = 0$. When $\lambda=1$, this recovers the standard PPI estimator; when $\lambda=0$, it reduces to pure imputation using predictions; intermediate values interpolate between these extremes. The parameter $\lambda$ can be chosen to optimize statistical power while maintaining validity, though the default choice $\lambda=1$ provides valid inference without tuning.

\paragraph{Example: population mean estimation.}
For estimating the population mean $\theta^* = \E[Y]$, the PPI estimator takes the form:
\begin{equation*}
\hat{\theta}_{\mathrm{PPI}} = \underbrace{\frac{1}{N}\sum_{i=1}^N f(\tilde{X}_i)}_{\text{prediction average}} + \underbrace{\frac{1}{n}\sum_{i=1}^n (Y_i - f(X_i))}_{\text{bias correction}}
\end{equation*}
Under the assumption that $f$ is independent of the inference data and both samples are i.i.d.\ from the same distribution, unbiasedness follows:
\begin{equation*}
\E[\hat{\theta}_{\mathrm{PPI}}] = \E[f(X)] + \E[Y - f(X)] = \E[Y] = \theta^*
\end{equation*}

\begin{table*}[htbp]
\centering
\small
\begin{tabular}{@{}llll@{}}
\toprule
\textbf{Estimand} & \textbf{Prediction-based score} $\hat g_\theta^{\mathrm{pred}}$ & \textbf{Rectifier} $\widehat{\Delta}_\theta$ & \textbf{Procedure} \\
\midrule
Mean & $\theta - \frac{1}{N}\sum_{i=1}^N f(\tilde X_i)$ 
     & $\frac{1}{n}\sum_{i=1}^n \big(f(X_i)-Y_i\big)$ 
     & Alg.~1 \\
\addlinespace
Median ($q=\tfrac{1}{2}$) 
     & $\tfrac{1}{2} - \frac{1}{N}\sum_{i=1}^N \mathbf{1}\{f(\tilde X_i)\le \theta\}$ 
     & $\frac{1}{n}\sum_{i=1}^n \big(\mathbf{1}\{f(X_i)\le \theta\}-\mathbf{1}\{Y_i\le \theta\}\big)$ 
     & Alg.~2 \\
\addlinespace
$q$-quantile 
     & $q - \frac{1}{N}\sum_{i=1}^N \mathbf{1}\{f(\tilde X_i)\le \theta\}$ 
     & $\frac{1}{n}\sum_{i=1}^n \big(\mathbf{1}\{f(X_i)\le \theta\}-\mathbf{1}\{Y_i\le \theta\}\big)$ 
     & Alg.~2 \\
\addlinespace
Logistic reg. 
     & $\frac{1}{N}\sum_{i=1}^N \tilde X_i\!\left(\sigma(\theta^\top \tilde X_i)- f(\tilde X_i)\right)$ 
     & $\frac{1}{n}\sum_{i=1}^n X_i\!\left(f(X_i)-Y_i\right)$ 
     & Alg.~3 \\
\addlinespace
Linear reg. 
     & $\frac{1}{N}\sum_{i=1}^N \big(\tilde X_i \tilde X_i^\top \theta - \tilde X_i f(\tilde X_i)\big)$ 
     & $\frac{1}{n}\sum_{i=1}^n X_i\!\left(f(X_i)-Y_i\right)$ 
     & Alg.~4 \\
\addlinespace
Convex minimizer 
     & $\frac{1}{N}\sum_{i=1}^N \nabla \ell_\theta(\tilde X_i, f(\tilde X_i))$ 
     & $\frac{1}{n}\sum_{i=1}^n \big(\nabla \ell_\theta(X_i, Y_i) - \nabla \ell_\theta(X_i, f(X_i))\big)$ 
     & Alg.~5 \\
\bottomrule
\end{tabular}
\caption{Prediction-powered estimating equations for common estimands. Here $\sigma$ denotes the logistic sigmoid.}
\label{tab:ppi_estimands}
\end{table*}


\subsection{Mapping \(\phi:\mathcal{Y}\to\mathbb{R}\)}
\label{app:mapping}
\paragraph{NHANES.}
We use the USDA FNDDS database to obtain nutrient profiles and define a mapping function $\phi$ that returns a \emph{scalar} target from open-ended text (e.g., energy in kcal for a food mention). For each simulated food item, we first retrieve the top 40 candidate foods by cosine similarity over FNDDS (using \texttt{snowflake-arctic-embed} \cite{yu2024arctic}). These candidates and the query are then passed to GPT-4o-mini with temperature 0 in a RAG-style setup \cite{lewis2020retrieval, russo2025meat} to select the best match and we retrieve its kcal density. Using the retrieved kcal density and the reported grams for each food item, we compute the total daily caloric intake. Importantly, this mapping function must remain deterministic so as not to introduce unnecessary variance into the correction step.

\paragraph{ATP.}
For ATP, $\phi$ maps ordinal multiple choice responses to numerical values. Specifically, we assign A$\mapsto$1, B$\mapsto$2, C$\mapsto$3, D$\mapsto$4, and so forth.

\subsection{Prompts}
\label{app:prompts}

\paragraph{NHANES.} The prompt for simulating responses for NHANES is included in the box below.

\paragraph{ATP.} The prompt for simulating responses for ATP questions is included in the boxes below.

\paragraph{\texttt{Persona-guided} prompt template.}
To generate detailed persona descriptions from demographic and survey response data, we used the following prompt template:

\begin{tcolorbox}[
  colback=white,
  colframe=black!50,
  coltitle=white,
  fonttitle=\bfseries,
  title=\textbf{NHANES prompt template},
  colbacktitle=black!50,
  boxrule=0.4mm,
  toptitle=1mm,
  bottomtitle=1mm,
]
\textbf{System message:}\\[1mm]
You are participating in a dietary recall survey. Describe what you ate and drank in the past 24 hours, based on your memory. Below is your personal profile:\\
\texttt{\{demographic\_profile\}}\\[1mm]
Answer honestly and realistically as you are recalling from memory.\\
\textbf{User prompt:}\\[1mm]
Please list everything you ate and drank in the past 24 hours. Include meals, snacks, drinks, and small bites, in the order you consumed them.\\[1mm]
Use this exact format on each line:\\
\texttt{[Food name] - [Short description] - [Grams as a number only]}\\[1mm]
Instructions:
\begin{itemize}[topsep=1pt,itemsep=1pt,parsep=0pt,partopsep=0pt]
  \item One item per line.
  \item Use a single hyphen and space (\texttt{" - "}) to separate fields.
  \item Grams must be a number only, no units like ``g'' or ``grams''.
  \item Do not add summaries or explanations—only the list.
\end{itemize}
\end{tcolorbox}

\begin{tcolorbox}[
  colback=white,
  colframe=black!50,
  coltitle=white,
  fonttitle=\bfseries,
  title=\textbf{ATP prompt template},
  colbacktitle=black!50,
  boxrule=0.4mm,
  toptitle=1mm,
  bottomtitle=1mm,
]
\textbf{System message:}
You are a survey respondent. Adopt this profile: \texttt{\{demographic\_profile\}} Answer as yourself in first person. Pick exactly one option from the list. Output only one uppercase letter from \{letter choices\}. No words, no punctuation, no explanations, no qualifiers. Do not discuss ethics, study design, or what most people would do.

\textbf{User prompt:}
\texttt{\{survey\_question\}}

Choices:
\texttt{\{multiple\_choice\_options\}}
\end{tcolorbox}




\begin{tcolorbox}[
  colback=white,
  colframe=black!50,
  coltitle=white,
  fonttitle=\bfseries,
  title=\textbf{\texttt{Persona-guided} prompt for synthesizing personas based on historical data},
  colbacktitle=black!50,
  boxrule=0.4mm,
  toptitle=1mm,
  bottomtitle=1mm,
]
\textbf{System message:}
You are an expert at synthesizing detailed persona descriptions from demographic and behavioral data. Your output must be a single, coherent paragraph.

\textbf{User prompt:}
Here is a person's demographic information and a log of what they answered in the past.

\textbf{Demographics}
\texttt{\{demo\}}

\textbf{History}
\texttt{\{context\}}

Your task:
Based on this information, write a single, detailed paragraph describing this person's habits, lifestyle, and personality in general. Your description should be a coherent narrative that synthesizes all the available evidence. Focus on inferring patterns, routines, and constraints that are supported by the provided information. Do not use lists, bullet points, or scores. Do not include extra explanations or reasoning. Just provide the narrative persona description.
\end{tcolorbox}





\section{Data and Computational Resources}
\subsection{Dataset Licensing}
Both datasets used in our study are publicly available. The U.S. National Health and Nutrition Examination Survey (NHANES) is produced by the National Center for Health Statistics and is in the public domain. The American Trends Panel (ATP) is released by the Pew Research Center for scholarly use under its data use terms. No special licenses or permissions were required for access or use of these datasets in our work.

\subsection{Computational Resources}
All experiments were conducted using models with approximately 7-8 billion parameters.
Training and evaluation were performed on a single NVIDIA A100 GPUs. The total training duration across all runs was approximately 3 days ($\approx$72 GPU hours). This includes all fine-tuning, evaluation, and validation steps.

\end{document}